\documentclass[runningheads]{llncs}
\usepackage[T1]{fontenc}
\usepackage{booktabs}
\usepackage{graphicx}
\usepackage{amsmath}
\usepackage{amsfonts}
\usepackage{cite}
\usepackage{makecell}
\usepackage{hyperref}
\usepackage{xurl}
\usepackage[table, svgnames, dvipsnames]{xcolor}

\usepackage[accsupp]{axessibility}

\usepackage{color}

\usepackage{tikz}
\usepackage{multirow}
\usepackage{multicol}
\definecolor{gold}{HTML}{FBF2D2}
\definecolor{silver}{HTML}{DDDDDD}
\definecolor{bronze}{HTML}{EED2B8}

\definecolor{goldD}{HTML}{D9AE13}
\definecolor{silverD}{HTML}{909090}
\definecolor{bronzeD}{HTML}{9A5F26}

\definecolor{catGreen}{HTML}{238763}
\definecolor{catBlue}{HTML}{1F70AE}

\definecolor{SuperSimpleNet_clr}{HTML}{a1c9f4}
\definecolor{SimpleNet_clr}{HTML}{377eb8}
\definecolor{BGAD_clr}{HTML}{4daf4a}
\definecolor{PRN_clr}{HTML}{984ea3}
\definecolor{SegDecNet_clr}{HTML}{ff7f00}
\definecolor{DRA_clr}{HTML}{ffff33}
\definecolor{AST_clr}{HTML}{66c2a5}
\definecolor{PatchCore_clr}{HTML}{fc8d62}
\definecolor{FastFlow_clr}{HTML}{8da0cb}
\definecolor{Draem_clr}{HTML}{e78ac3}
\definecolor{DSR_clr}{HTML}{a6d854}
\definecolor{EfficientAD_clr}{HTML}{ffd92f}
	
\usepackage{color, colortbl}

\definecolor{sup_mask_gen}{HTML}{82B366}

\newcommand{\medal}[3]{\tikz[baseline=(char.base)]{\node[rounded corners=2pt,fill=#1,draw=#2,inner sep=1.5pt] (char) {#3};}}

\newcommand{\bm}[2]{
    \ifcase#1\or
      {\medal{gold}{goldD}{\textbf{#2}}}
    \or 
      {\medal{silver}{silverD}{#2}}
    \or 
      {\medal{bronze}{bronzeD}{#2}}
    \else 
      #2
    \fi\ignorespaces
}

\let\titleold\title
\renewcommand{\title}[1]{\titleold{#1}\newcommand{\thetitle}{#1}}
\def\maketitlesupplementary
   {
   \newpage
       {
        \centering
        \Large
        \textbf{\thetitle}\\
        \vspace{0.5em}Supplementary Material \\
        \vspace{1.0em}
       }
   }

\usepackage{xcolor}

\title{SuperSimpleNet: Unifying Unsupervised and Supervised Learning for Fast and Reliable Surface Defect Detection} 
\titlerunning{SuperSimpleNet}
\author{Bla\v{z} Rolih \and
Matic Fu\v{c}ka \and
Danijel Sko\v{c}aj}
\authorrunning{B. Rolih et al.}
\institute{University of Ljubljana, Faculty of Computer and Information Science, Slovenia
\email{br9136@student.uni-lj.si, \{matic.fucka, danijel.skocaj\}@fri.uni-lj.si}\\
}

\begin{document}
\maketitle              
\begin{abstract}
The aim of surface defect detection is to identify and localise abnormal regions on the surfaces of captured objects, a task that's increasingly demanded across various industries. Current approaches frequently fail to fulfil the extensive demands of these industries, which encompass high performance, consistency, and fast operation, along with the capacity to leverage the entirety of the available training data. Addressing these gaps, we introduce SuperSimpleNet, an innovative discriminative model that evolved from SimpleNet. This advanced model significantly enhances its predecessor's training consistency, inference time, as well as detection performance. SuperSimpleNet operates in an unsupervised manner using only normal training images but also benefits from labelled abnormal training images when they are available. SuperSimpleNet achieves state-of-the-art results in both the supervised and the unsupervised settings, as demonstrated by experiments across four challenging benchmark datasets. Code: \href{https://github.com/blaz-r/SuperSimpleNet}{\color{magenta}{https://github.com/blaz-r/SuperSimpleNet}}.

\keywords{Surface Defect Detection \and Surface Anomaly Detection \and Industrial Inspection \and Supervised Learning \and Unsupervised Learning.}
\end{abstract}
\section{Introduction}

\begin{figure}[h]
  \centering
   \includegraphics[width=1\linewidth]{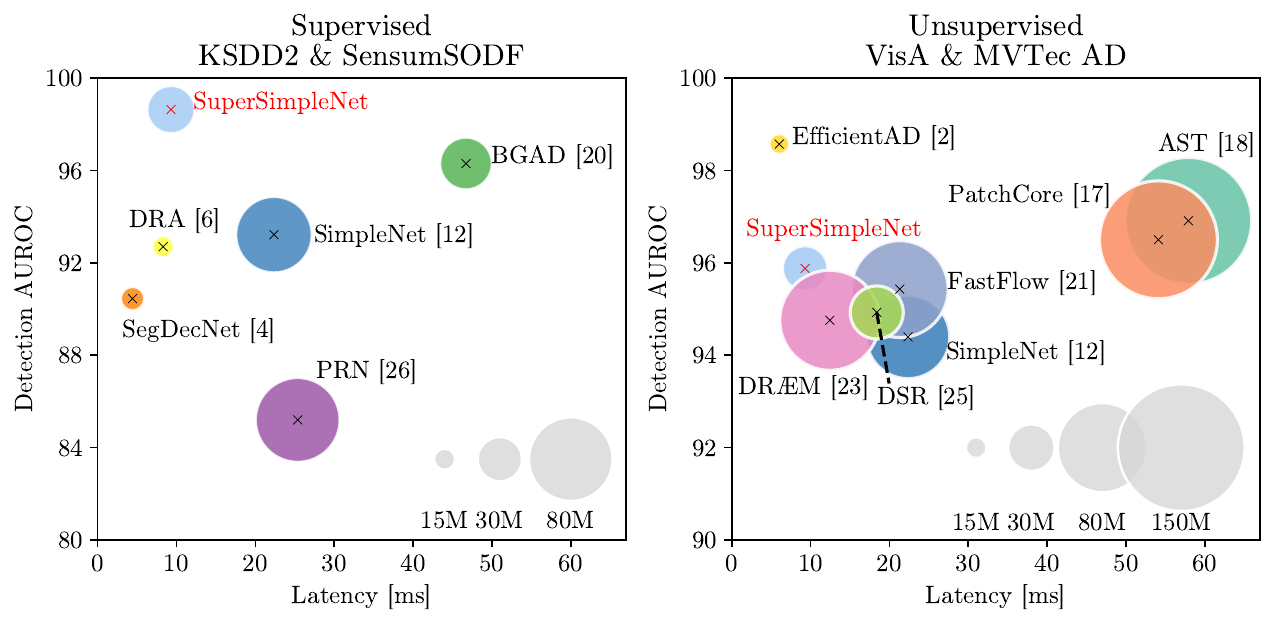}
    \begin{tabular}{l|ccccccccccccc|cc} 
\toprule Requirement & 
\makecell{
{\hypersetup{citebordercolor=AST_clr}
\cite{rudolph_ast}}
}
&\makecell{{\hypersetup{citebordercolor=FastFlow_clr}\cite{yu_fastflow}} } & \makecell{{\hypersetup{citebordercolor=PatchCore_clr}\cite{roth_patchcore}} } & \makecell{{\hypersetup{citebordercolor=Draem_clr}\cite{zavrtanik_draem}} } & \makecell{{\hypersetup{citebordercolor=PRN_clr}\cite{zhang_prn}} }&  \makecell{\cite{racki_sensum}} &  \makecell{\cite{ferrite_defect}} &  \makecell{\cite{luo_maminet}}  & \makecell{{\hypersetup{citebordercolor=DRA_clr}\cite{ding_dra}}  } & \makecell{{\hypersetup{citebordercolor=SegDecNet_clr}\cite{KSDD2}} } & \makecell{{\hypersetup{citebordercolor=EfficientAD_clr}\cite{batzner_efficientad}} } & \makecell{{\hypersetup{citebordercolor=BGAD_clr}\cite{yao_bgad}} } & \makecell{{\hypersetup{citebordercolor=DSR_clr}\cite{zavrtanik2022dsr}} } & \makecell{{\hypersetup{citebordercolor=SimpleNet_clr}\cite{liu_simplenet}}}&  \makecell{\textbf{\textcolor{red}{Ours}}}\\ \hline
Unsupervised& \checkmark & \checkmark & \checkmark & \checkmark & ~ & ~ & ~ & ~ & ~ & ~ & \checkmark & * & \checkmark & \checkmark & \checkmark \\
Supervised& ~ & ~ & ~ & ~ & \checkmark & \checkmark & \checkmark & \checkmark & \checkmark & \checkmark & ~ & \checkmark & * & * & \checkmark \\
Speed& ~ & ~ & ~ & ~ & ~ & \checkmark & - & - & \checkmark & \checkmark & \checkmark & ~ & ~ & ~ & \checkmark\\
\bottomrule
    \end{tabular}
   \caption{Model comparison for both the supervised (KSDD2~\cite{KSDD2} and SensumSODF~\cite{racki_sensum}) and the unsupervised (MVTec AD~\cite{mvtec} and VisA~\cite{visa}) setting. The Y-axis represents the anomaly detection performance measured in AUROC, and the X-axis represents inference time in milliseconds using an NVIDIA Tesla V100S (more details in Section~\ref{sec:results}). The size of the circles represents the model's parameter size. Additionally, the table below indicates whether each model meets specific speed requirements (if its inference time is below 10ms) and whether it is capable of working in the unsupervised and/or the supervised setting. If a model is designed specifically for either the supervised or the unsupervised setting but theoretically applicable to the other, we marked the opposing cell with a '*'. Two methods (marked with '-') lack publicly available code, preventing us from assessing their speed. SuperSimpleNet stands out as the only model meeting all criteria.} 
   \label{fig:model_comp}
   \vspace{-1px}
\end{figure}

A critical aim of the manufacturing process is to achieve high-quality products and increased efficiency. Surface Defect Detection (SDD) plays a pivotal role in this pursuit, as it aims to identify and classify defects or irregularities on the surface of manufactured components. Traditional manual inspection methods are time-consuming, subjective, and prone to human error. In contrast, automated SDD systems offer the potential for real-time monitoring, precise defect localisation, and improved product quality. The integration of deep learning algorithms into SDD systems~\cite{KSDD2, racki_sensum, mvtec, visa} has shown promising results, indicating their potential to revolutionise quality control processes and streamline manufacturing operations.

To bridge the gap between academic research and real-world manufacturing processes, developed models must meet all industry-defined requirements. These requirements fall into two main categories: performance and flexibility. Performance requirements pertain to the model's anomaly detection efficacy and its inference time. The model should exhibit a high anomaly detection rate alongside rapid inference capabilities. Although the performance aspects of these requirements have been extensively studied~\cite{batzner_efficientad, liu_simplenet}, flexibility requirements have largely been overlooked. Flexibility requirements are concerned with the model's adaptability to various training regimes encountered in actual manufacturing settings.
Different levels of annotations are available for various objects during training, requiring the model to be capable of using all available data effectively. This implies the necessity for a model to be trainable in both the supervised and the unsupervised setting, a feature seldom found in existing methods~\cite{zavrtanik2022dsr, yao_bgad}.
Another important yet often overlooked requirement is the stability of training, which should lead to consistent detection results, regardless of the specific training run. Unfortunately, a high level of consistency is not met by many existing methods. Our objective was to devise a method that fully meets these criteria: (i) detection and localisation performance, (ii) low inference time in both (iii) the supervised and the unsupervised setting, and (iv) a stable and consistent learning process. 

To meet all specified requirements, we introduce SuperSimpleNet, an innovative model that builds on the foundation laid by SimpleNet~\cite{liu_simplenet}. While SimpleNet has demonstrated great performance in the unsupervised setting, achieving these results consistently requires multiple training runs. This limitation, coupled with the industry's demand for efficient and more resilient models, prompted us to refine the training approach and architecture of the original model. These enhancements have rendered it more robust and suitable for practical applications. 

The main contributions of our work are as follows:
\begin{itemize}
    \item We propose SuperSimpleNet - a strong discriminative defect detection model tailored to meet industry standards. We optimised the originally proposed architecture and introduced a novel synthetic anomaly generation procedure, resulting in a more stable learning process and improved performance. With an inference time of 9.3 ms and a throughput of 268 images per second, SuperSimpleNet outspeeds most contemporary models while achieving state-of-the-art defect detection results. 
    \item We have extended the architecture, initially designed for the unsupervised setting, to incorporate abnormal training images and utilise available labels. Additionally, we've integrated a separate classification head into the model, which helps the model consider the image's global context. This unification of unsupervised and supervised approaches significantly boosts anomaly detection capabilities, positioning SuperSimpleNet among the top performers in both unsupervised and supervised scenarios. This versatility renders it highly suitable for industrial applications.     
\end{itemize}

We have performed extensive experiments on four challenging datasets. First, we show that SuperSimpleNet achieves state-of-the-art results in the supervised setting on two standard defect detection datasets -- SensumSODF and KSDD2, with an AUROC of 97.8\% and a detection AP of 97.4\%, respectively. Then, we show that SuperSimpleNet achieves state-of-the-art results in the unsupervised setting on two standard anomaly detection datasets -- MVTec AD and VisA, with an AUROC of 98.4\% and 93.4\%, respectively. With the state-of-the-art results achieved in both scenarios, as depicted in Fig~\ref{fig:model_comp}, we demonstrate the versatility of SuperSimpleNet and its suitability for real-life scenarios.

\section{Related Work}

\noindent\textbf{Unsupervised methods}

\noindent Unsupervised methods have become a cornerstone in surface anomaly detection because they can detect outliers in settings where labelled defect data is scarce or non-existent. To address such scenarios, various paradigms of models emerged. The \textit{reconstructive approaches} made the first attempts, which train an autoencoder-like network~\cite{zavrtanik_riad} with the idea that the model will successfully reconstruct only the anomalous regions whilst leaving the normal regions intact. Methods under this paradigm also used different kinds of generative networks, such as GANs~\cite{akcay_ganomaly} or transformers~\cite{intra}. The successful reconstruction of anomalous regions does not always hold, leading to an overall bad performance.

Another prominent paradigm involves \textit{leveraging the features} extracted from a pretrained network, such as a ResNet~\cite{he_resnet}. The extracted features are then used to learn normality by utilising approaches such as a normalising flow~\cite{rudolph_ast, yu_fastflow}, a memory-bank~\cite{roth_patchcore}, a student-teacher architecture~\cite{zhang_destseg} or distillation~\cite{reverse_dist}.

The remaining set of approaches, the \textit{discriminative methods}, are trained using synthetic anomalies. These anomalies can be generated on images~\cite{zhang_destseg, zavrtanik_draem, fuvcka_transfusion}, while the recent SimpleNet~\cite{liu_simplenet} generates them in latent space by perturbing the entire copy of features with noise. Another, more sophisticated approach to generating synthetic anomalies in the latent space is introduced in DSR~\cite{zavrtanik2022dsr}, where a Perlin noise mask conditions the area of anomalies.

While these methods have demonstrated success, their complexity in design and implementation often hinders efficient execution, leading recent research efforts to also prioritise efficiency~\cite{batzner_efficientad}. Even though these methods show great results in the unsupervised setting, most of them lag behind on datasets~\cite{KSDD2} curated for the supervised setting, as they cannot utilise labelled defects during training.

\noindent\textbf{Supervised methods}

\noindent Although defective samples are initially rare, their availability in industrial settings increases over time. However, the unsupervised methods often aren't designed to utilise such data effectively. Consequently, supervised anomaly detection is prevalent in industrial settings to maximise performance.
Industrial-grade methods such as SegDecNet~\cite{KSDD2}, TriNet~\cite{racki_sensum}, and MaMiNet~\cite{luo_maminet} leverage both normal and anomalous samples, with the capability to incorporate image-level labels in a weakly-supervised setting, thus alleviating the work of manual pixel-precise annotation. A downside of this approach is that anomalous data is still needed, albeit not annotated, and presents quite a different paradigm from unsupervised learning. 

There have been attempts to use one-class classification methods for anomaly detection, enabling unsupervised and supervised training. However, methods such as Deep SAD~\cite{ruff_deepSAD} and FCDD~\cite{liznerski_explainableOCC} demonstrate poor performance compared to recent methods, especially in the unsupervised setting. 

Other recent approaches, such as BGAD~\cite{yao_bgad}, PRN~\cite{zhang_prn}, and DRA~\cite{ding_dra}, have extended the supervised approach with synthetic anomaly generation. 
The main disadvantage of such methods is that they require many known anomalous samples to outperform recent unsupervised methods, and their performance in a strictly unsupervised setting may be inadequate.
Furthermore, these methods generate anomalies at the image level, although recent advancements in data augmentation could benefit from latent space generation. Additionally, the complexity of architectures, e.g. BGAD and PRN, leads to longer inference times, potentially limiting their applicability in industrial settings. 

\section{SuperSimpleNet}

\begin{figure}[t]
    \centering
    \includegraphics[width=1\linewidth]{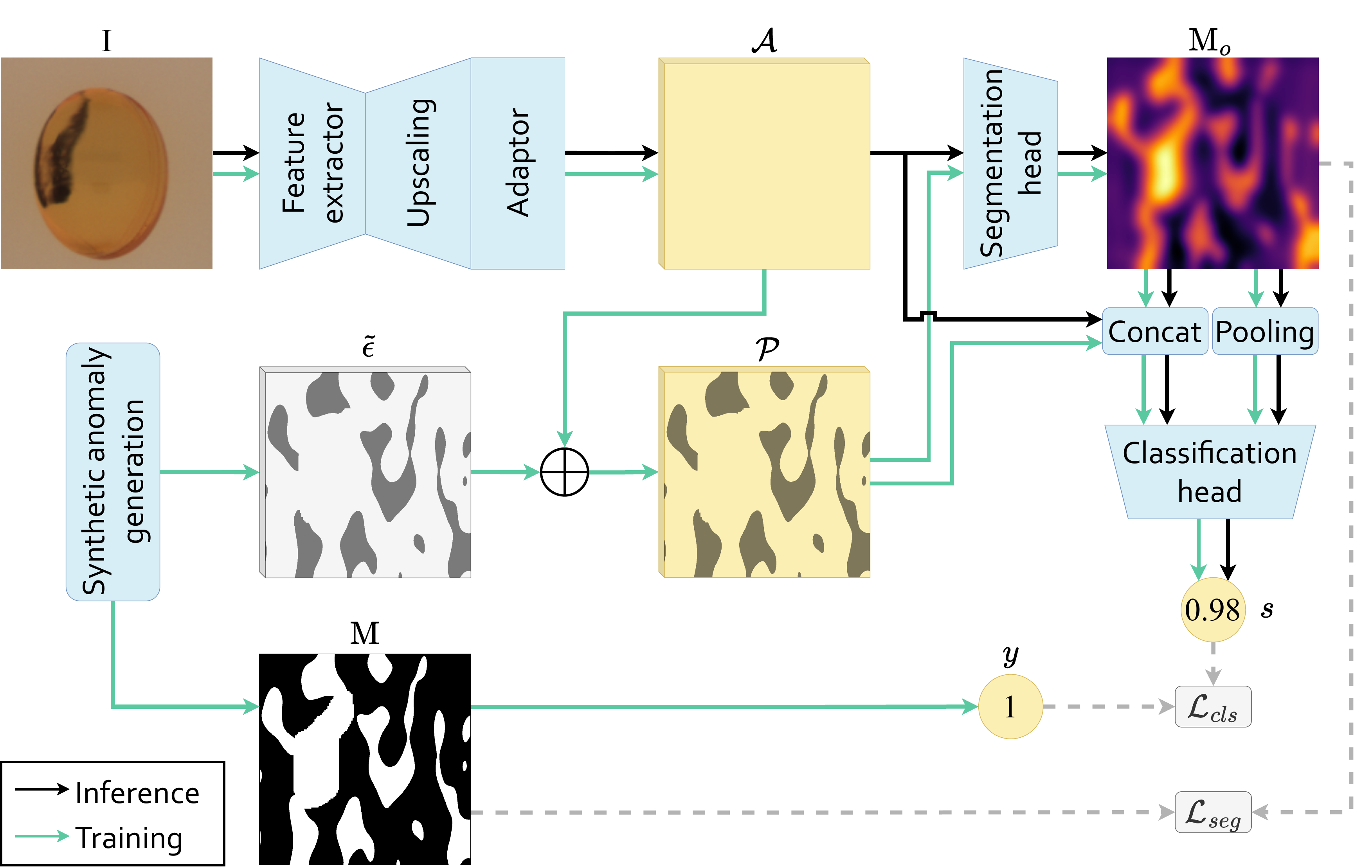}
    \caption{
    SuperSimpleNet's architecture. Features are first extracted, upscaled, and adapted. During training, synthetic anomalies are generated in latent space by adding Gaussian noise to the adapted feature map $\mathcal{A}$. The noise is limited to regions generated by binarised Perlin mask and non-anomalous regions (depicted by $\tilde{\epsilon}$). The perturbed feature map $\mathcal{P}$ is then used as the input for the segmentation head to predict an anomaly mask $\mathrm{M}_o$. The predicted anomaly mask $\mathrm{M}_o$ and the perturbed feature map $\mathcal{P}$ are then used as the input for the classification head, producing the anomaly score $s$. The produced anomaly score $s$ and the predicted mask $\mathrm{M}_o$ are during the training supervised by the anomaly mask $\mathrm{M}$ and $y$, where $y$ is set to 1 if the image contains an anomaly (synthetic or real) and to 0 otherwise. During inference, the anomaly generation phase is omitted, and $\mathrm{M}_o$ and $s$ are produced directly from the adapted feature map $\mathcal{A}$.
    }
    \label{fig:arch}
\end{figure}

The proposed SuperSimpleNet, as illustrated in Figure~\ref{fig:arch}, builds upon the foundation laid by SimpleNet~\cite{liu_simplenet}. It begins with feature extraction via a pretrained convolutional network (detailed in Section~\ref{sec:feat_extr}), followed by upscaling and pooling processes designed to encapsulate the neighbouring context. Subsequently, these features are adapted to a common latent space via a feature adaptor (further described in Section~\ref{sec:feat_adapt}). A notable enhancement over the original model is the introduction of an innovative approach for generating synthetic anomalies, playing a pivotal role in the model's enhanced performance across both unsupervised and supervised scenarios. This advancement is primarily attributed to the creation of anomaly regions at the feature level, employing a binarised Perlin noise mask (expounded in Section~\ref{sec:ano_gen}). The refined features are then funnelled into the segmentation and classification modules (outlined in Section~\ref{sec:seg_dec}). 

This method exclusively depends on the synthetically produced masks and labels in the unsupervised setting. However, the inclusion of synthetic anomalies markedly elevates the system's efficacy also in the supervised setting, particularly when integrated with actual ground truth data (discussed in Section~\ref{sec:ano_gen} and Section~\ref{sec:loss}). During inference, the framework operates in a seamless end-to-end fashion (Section~\ref{sec:inference}).

Subsequent sections will provide a thorough exposition of each module, encompassing training and inference details, ensuring a comprehensive understanding of the system's architecture and functionality.

\subsection{Feature extractor}
\label{sec:feat_extr}

Following the design principles of SimpleNet~\cite{liu_simplenet}, we employ a ResNet-like~\cite{he_resnet} convolutional neural network pretrained on ImageNet as the feature extractor. Specifically, we utilise a WideResNet50~\cite{zagoruyko_wideresnet}, extracting features from its 2nd and 3rd layers. 
Due to ResNet-like networks' architectural characteristics, output features are of relatively low resolution. This limits the effective detection of smaller anomalies and compromises the segmentation precision.

To effectively address this challenge, we have extended the base SimpleNet extractor by incorporating a new upscaling strategy prior to feature concatenation. Our methodology introduces an additional layer of upscaling, effectively doubling the previously applied scaling factor. As a result, layer 3 is enlarged by a factor of 4, while layer 2 undergoes a doubling in size. This approach ensures both layers achieve equal dimensions, thus allowing for a seamless concatenation. 

Afterwards, as in SimpleNet~\cite{liu_simplenet}, the neighbouring context is encapsulated through the application of local average pooling, employing a mean kernel of size \(3\times3\). This step yields an upscaled feature map where every element is enriched with information from its surroundings.
\subsection{Feature adaptor}
\label{sec:feat_adapt}

While the representations from pretrained backbones can transfer well to the task of anomaly detection~\cite{heckler_feature_importance}, a feature adaptor as in SimpleNet~\cite{liu_simplenet} is employed to further improve the features for the task. It is implemented as a simple linear layer, producing the adapted features, denoted as $\mathcal{A}$.

\subsection{Feature-space anomaly generation}
\label{sec:ano_gen}

\begin{figure}[t]
    \centering
    \includegraphics[width=1\linewidth]{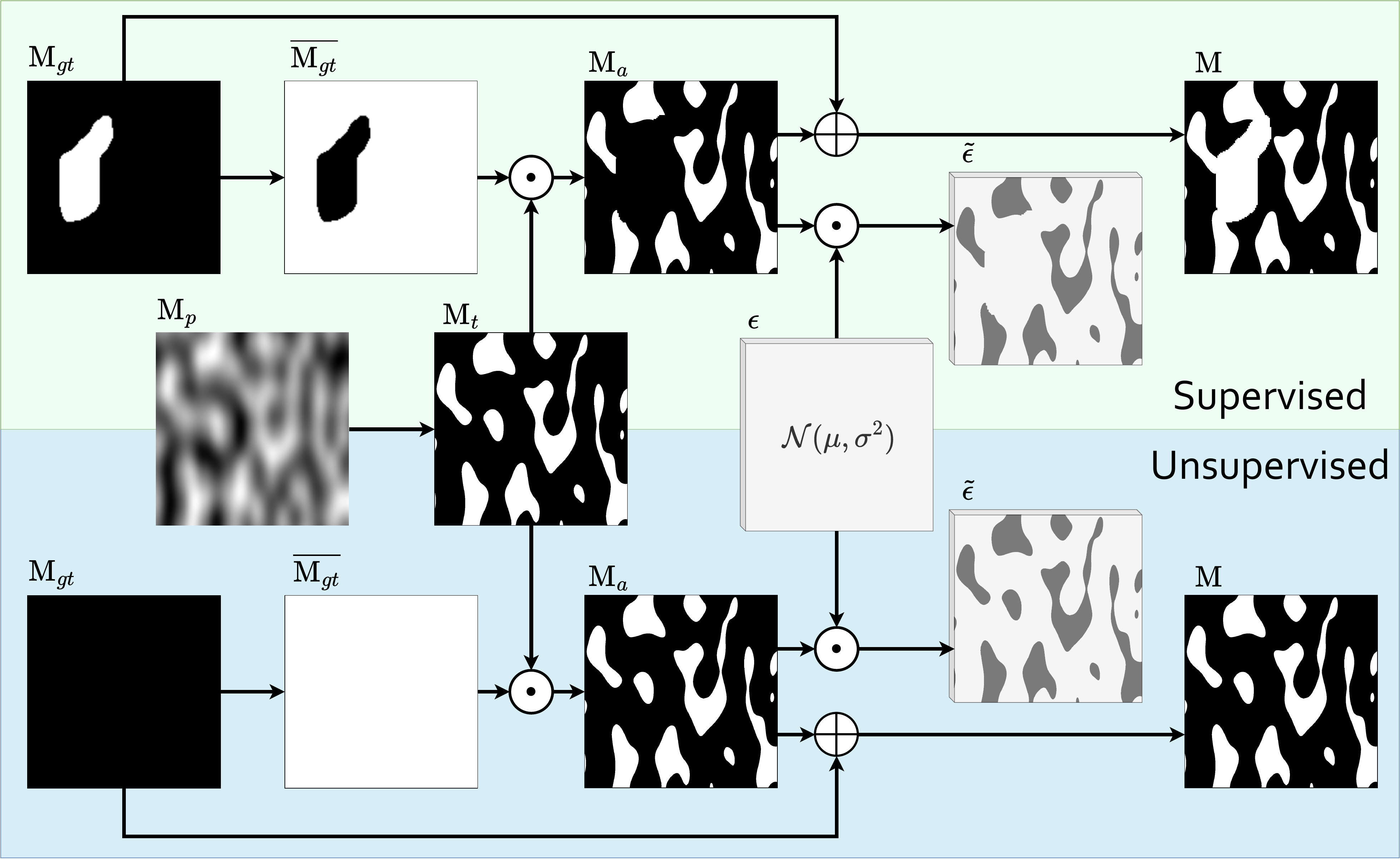}
    \caption{Synthetic anomaly generation. Synthetic anomaly masks $\mathrm{M}_a$ are generated using Perlin Noise. In the unsupervised setting, Gaussian noise is added to all the regions denoted by the thresholded Perlin Noise mask $\mathrm{M}_t$. In contrast, in the supervised setting, noise is omitted from the regions with actual anomalies, denoted by $\mathrm{M}_{gt}$. The final anomaly mask $\mathrm{M}$ is constructed from $\mathrm{M}_a$ and $\mathrm{M}_{gt}$, and holds information on both where the Gaussian Noise is added and where the actual anomalies lie.}
    \label{fig:noise}
\end{figure}

The anomaly generation process (visualised in Figure~\ref{fig:noise}) begins with the creation of a Perlin binary anomaly mask $\mathrm{M}_t$ by thresholding a generated Perlin noise image $\mathrm{M}_p$ (similarly as in~\cite{zavrtanik_draem, fuvcka_transfusion, zhang_destseg}). All the regions containing actual anomalies, delineated by $M_{gt}$, are removed from $M_t$, resulting in $M_a$ (Figure~\ref{fig:noise} -- \textcolor{sup_mask_gen}{green} section). In the unsupervised setting, $M_{gt}$ is always empty during training (Figure~\ref{fig:noise} -- \textcolor{blue}{blue} section). Subsequently, Gaussian noise $\epsilon$, sampled from the Gaussian distribution $\mathcal{N}(\mu, \sigma^2)$, is selectively applied only within the confines defined by $\mathrm{M}_a$, as illustrated in Figure~\ref{fig:noise}, and then added to the adapted features $\mathcal{A}$ to produce the perturbed feature map $\mathcal{P}$. To increase the stability of the training regime, the duplication of the adapted features is retained, but unlike the original SimpleNet, noise is applied to both the original and the copy following the process described above. Through this refined strategy, SuperSimpleNet achieves a higher level of precision in anomaly simulation, focusing on creating more realistic, spatially coherent, yet highly randomised anomalous regions. This inherent randomness safeguards the model against overly depending on specific patterns, which might not be representative of unseen data.
 
A direct method for the \textit{supervised learning setting} might involve substituting the synthetically generated anomaly mask, $\mathrm{M}_{a}$, with the actual ground truth anomaly mask, $\mathrm{M}_{gt}$. Yet, we recognised that the defects present in the training dataset often fail to fully encompass the vast spectrum of potential defects. To address this limitation, we supplement the model with additional synthetic anomalies created using the methodology employed in the unsupervised framework to represent the defect distribution more comprehensively. Gaussian noise is added solely to non-defective areas to ensure the model gets as much actual defect data as possible. Consequently, the anomaly masks $\mathrm{M}$ for the supervised learning scenario are formulated from a blend of authentic ground truth data and synthetic anomalies, enriching the model's exposure to a wider array of defect variations and bolstering its detection and generalisation capabilities.

\subsection{Segmentation-detection module}
\label{sec:seg_dec}

We have further extended the architecture to enhance anomaly detection performance by introducing a \textit{classification head}, $D_{cls}$, while maintaining the \textit{segmentation head}, $D_{seg}$, as it was in SimpleNet. The classification head's design is straightforward, consisting of a single $5 \times 5$ convolutional block and a linear layer. This structure allows the model to understand the global semantics of the image, reducing the overall count of false positives. It also makes it easier for the model to detect significant changes in small areas, thereby increasing the detection rate of smaller anomalies that might have gone unnoticed before.

As displayed in Figure~\ref{fig:arch}, an anomaly mask $\mathrm{M}_o$ is first produced using the segmentation head. This mask $\mathrm{M}_o$ is then concatenated with the adapted feature map $\mathcal{A}$ (or noise-augmented feature map $\mathcal{P}$ during training) and used as input for the convolutional block of the classification head. Both, the output from the convolutional block and the anomaly map $\mathrm{M}_o$ undergo average pooling and max pooling, after which they are combined and fed into the final linear layer, producing an image-level anomaly score $s$.

\subsection{Loss function}
\label{sec:loss}

The truncated $\mathcal{L}_1$ loss is used for the segmentation head. The loss consists of two cases defined in Equation~\ref{eq:l1_parts} where $th$ is the truncation term preventing overfitting (in our case, 0.5): 
\begin{equation}
    l_{i,j} = 
    \begin{cases}
        max(0, th - D_{seg}(P_{i, j})) \text{; if } \mathrm{M[i,j]} = 1\\
        max(0, th + D_{seg}(P_{i, j})) \text{; otherwise}
    \end{cases}
    \enspace .
    \label{eq:l1_parts}
\end{equation}
The total truncated $\mathcal{L}_1$ loss, denoted by $\mathcal{L}_{1t}$, is computed by averaging the loss $l_{i,j}$ across all elements within the predicted anomaly mask.
To better guide and stabilise the training process, we additionally incorporate focal loss~\cite{lin_focal} due to its improved performance in scenarios with unbalanced data, resulting in the formulation of the final segmentation loss, denoted as $\mathcal{L}_{seg}$:
\begin{equation}
    \mathcal{L}_{seg} = \mathcal{L}_{1t} + \mathcal{L}_{foc} \enspace .
    \label{eq:seg_loss}
\end{equation}
\noindent For the classification loss $\mathcal{L}_{cls}$, the focal loss~\cite{lin_focal} is employed:
\begin{equation}
    \mathcal{L}_{cls} = \mathcal{L}_{foc} \enspace .
    \label{eq:cls_loss}
\end{equation}
The final loss is the sum of the segmentation and classification loss:
\begin{equation}
    \mathcal{L} = \mathcal{L}_{seg} + \mathcal{L}_{cls} \enspace .
    \label{eq:total_loss}
\end{equation}
The target in segmentation loss is the anomaly mask $\mathrm{M}$, which delineates the regions with synthetic and real anomalies. The target anomaly label $y$ for classification loss is derived from $\mathrm{M}$, i.e. $y$ is set to 1 if the image contains an anomaly (synthetic or real) and to 0 otherwise.

\subsection{Inference}
\label{sec:inference}

During inference, the network predicts the anomaly map and anomaly score by bypassing the anomaly generation phase. The segmentation head outputs an anomaly mask $\mathrm{M}_o$. This mask undergoes interpolation to match the input image's size and is further refined by applying a Gaussian filter with $\sigma=4$, yielding the final anomaly map. The anomaly score for each image is given by the value $s$, produced by the classification head $D_{cls}$.

\section{Experiments}

An extensive evaluation of the proposed method is performed in both the supervised and the unsupervised setting. The setup is described first, followed by the results.

\subsection{Datasets}
\label{subsec:datasets}

The performance in the supervised setting is evaluated on two real-world, reliable, and well-annotated datasets: Sensum Solid Oral Dosage Forms (SensumSODF)~\cite{racki_sensum} and Kolektor Surface-Defect Dataset 2 (KSDD2)~\cite{KSDD2}. 
SensumSODF consists of two categories, each a different type of solid oral dosage form: a capsule and a softgel. Both categories contain normal and annotated anomalous samples with defects of varying complexity and size. SensumSODF does not contain a predefined train-test split. Due to that, we followed an already defined protocol~\cite{racki_sensum}, involving a 3-fold cross-validation.
KSDD2 is constructed from images of production items captured using a visual inspection system. Both train and test split contain normal and precisely annotated anomalous samples with many in-distribution defects. Figure~\ref{fig:sup_qual} shows some examples from both datasets.

The unsupervised regime is evaluated on two established datasets: MVTec AD~\cite{mvtec} and VisA~\cite{visa}. MVTec AD encompasses 15 categories, while VisA comprises 12 different categories. Each category consists of a training set with only normal images and a test set with normal and pixel-precise annotated anomalous images. Anomalies present in both datasets are of various types, shapes, and scales. Figure~\ref{fig:unsup_qual} shows some examples from both datasets.

\subsection{Evaluation metrics}

Evaluation metrics depend on the used dataset and recent literature. Image-level performance for SensumSODF, MV\-Tec AD, and VisA is evaluated using the Area Under the Receiver Operator Curve (AUROC). 
Recent works have strayed away from using pixel-level AUROC for the pixel-level evaluation on these three datasets and rather opted for Area Under the Per-Region Overlap (AUPRO). We chose the same. In the case of KSDD2, a vast majority of recent works evaluate the image and pixel-level performance using Average Precision ($AP_{det}$ and $AP_{loc}$). Once again, we followed the suite of previous works. 

\subsection{Implementation details}
\label{exp:impl}

The model is trained for 300 epochs with a batch size of 32 using the AdamW optimiser. The adaptor module has a learning rate set to $10^{-4}$, while both the segmentation and classification head have it set to $2*10^{-4}$ with a weight decay of $10^{-5}$. To improve the inference time of the feature adaptor and the segmentation head, we exchange the feature vector reshaping and the consequent fully connected layers with a simple $1\times1$ convolution, which yields the same results. 

The learning rate scheduler is utilised to enhance training stability, multiplying the learning rate by 0.4 after 240 and 270 epochs. To further stabilise the training, the gradient is adjusted by stopping the gradient flow from the classification head to the segmentation head in the unsupervised setting and clipping the gradient in the supervised setting to norm 1.

Following SimpleNet~\cite{liu_simplenet}, Gaussian noise is sampled from $\mathcal{N}(0, \sigma^2)$ with $\sigma=0.015$. Perlin noise is binarised using a threshold of 0.6 for VisA, KSDD2, and SensumSODF, generating thinner and smaller anomalies. Since MVTec AD contains larger anomalies, a threshold of 0.2 is used. Synthetic anomalies are added to $50\%$ of the images.

All input images are normalised using ImageNet normalisation. MVTec AD and VisA use image dimensions of $256 \times 256$ without center-crop. For KSDD2, following the original protocol, a $232 \times 640$ resolution is used. For SensumSODF, capsule and softgel categories have resolutions of $192 \times 320$ and $144 \times 144$, respectively.
To extend the anomalous samples, flipping is used as in~\cite{racki_sensum} where batches are composed of equal anomalous and normal samples, as in~\cite{KSDD2}. As stated in Subsection~\ref{subsec:datasets}, we followed predefined train-test splits for KSDD2~\cite{KSDD2}, MVTec AD~\cite{mvtec}, and VisA~\cite{visa}, while we used 3-fold cross-validation for SensumSODF, as defined in the original paper~\cite{racki_sensum}.

For comparison, we extended the original SimpleNet to support training in a supervised manner. 
This was done by changing the loss design to consider the labelled defects, i.e. predicting the defective regions inside the ground truth mask ($\mathrm{M}_{gt}$) as anomalous.
All other parameters are kept the same as the original~\cite{liu_simplenet}.

The metrics are calculated based on the model resulting from the final epoch in both settings and all categories. For SuperSimpleNet and SimpleNet, the average performance of 5 runs with different seeds is reported, along with the corresponding standard deviations.

\subsection{Experimental results}
\label{sec:results}

\noindent\textbf{Results in the supervised setting.} 
We compared our method with the current state-of-the-art methods for the supervised setting: SegDecNet~\cite{KSDD2}, DRA~\cite{ding_dra}, BGAD~\cite{yao_bgad}, PRN~\cite{zhang_prn}, TriNet~\cite{racki_sensum} and SimpleNet~\cite{liu_simplenet}. The results on SensumSODF are displayed in Table~\ref{tab:sensum}, where SuperSimpleNet achieves the best result with a mean anomaly detection AUROC of $97.8\%$, surpassing the previous state-of-the-art by $0.9$ percentage points (p.p.), reducing the error by 29\%. Anomaly detection and localisation results in a supervised setting on KSDD2 are shown in Table~\ref{tab:ksdd2}. SuperSimpleNet achieves a state-of-the-art $\text{AP}_{\text{det}}$ of $97.4\%$. We hypothesise that SuperSimpleNet achieves such a high performance in the supervised setting due to the classification head, which can efficiently learn to capture more global information due to the presence of real and synthetic anomalies during training.
\vspace{-3mm}
\begin{table}[!ht]
    \centering
    \setlength{\tabcolsep}{4pt}
    \begin{tabular}{lccccccc}
    \toprule
		~  & \cite{KSDD2}  & \cite{ding_dra}  & \cite{yao_bgad}  & \cite{zhang_prn}  & \cite{racki_sensum}  & \cite{liu_simplenet}  & Ours \\ \midrule
		Detection & 83.4& 90.1& 94.3& 80.6& 96.9& 88.4 $(\pm$ 1.84) & 97.8 $(\pm$ 0.13) \\
		  Localisation & 75.2& - & 97.0& 66.0& -& 89.6 $(\pm$ 1.14) & 93.0 $(\pm$ 0.54) \\
    \bottomrule
    \end{tabular}
    \caption{Results of Supervised anomaly detection (AUROC) and localisation (AUPRO) on the SensumSODF dataset.}
    \label{tab:sensum}
\end{table}
\vspace{-13mm}
\begin{table}[!h]
    \centering
    \setlength{\tabcolsep}{3pt}
        \begin{tabular}{lccccccccc}
        \toprule
		~  & \cite{zavrtanik2022dsr}  & \cite{KSDD2}  & \cite{ding_dra}  & \cite{yao_bgad}  & \cite{zhang_prn}  & \cite{luo_maminet}  & \cite{ferrite_defect}  & \cite{liu_simplenet}  & Ours \\ \midrule
		Detection & 95.2& 95.4& 89.3& 92.7& 78.6& 96.2& 99.9& 93.5 $(\pm$ 1.05) & 97.4 $(\pm$ 0.25) \\
		Localisation & 85.5& 67.6& - & 76.5& 48.2& - & - & 75.9 $(\pm$ 2.40) & 82.1 $(\pm$ 0.50) \\
        \bottomrule
    \end{tabular}
    \caption{Results for supervised anomaly detection ($\text{AP}_{\text{det}}$) and localisation ($\text{AP}_{\text{loc}}$) on the KSDD2 dataset.}
    \label{tab:ksdd2}
\end{table}
\vspace{-12mm}
\begin{figure}[!h]
  \centering
   \includegraphics[width=1\linewidth]{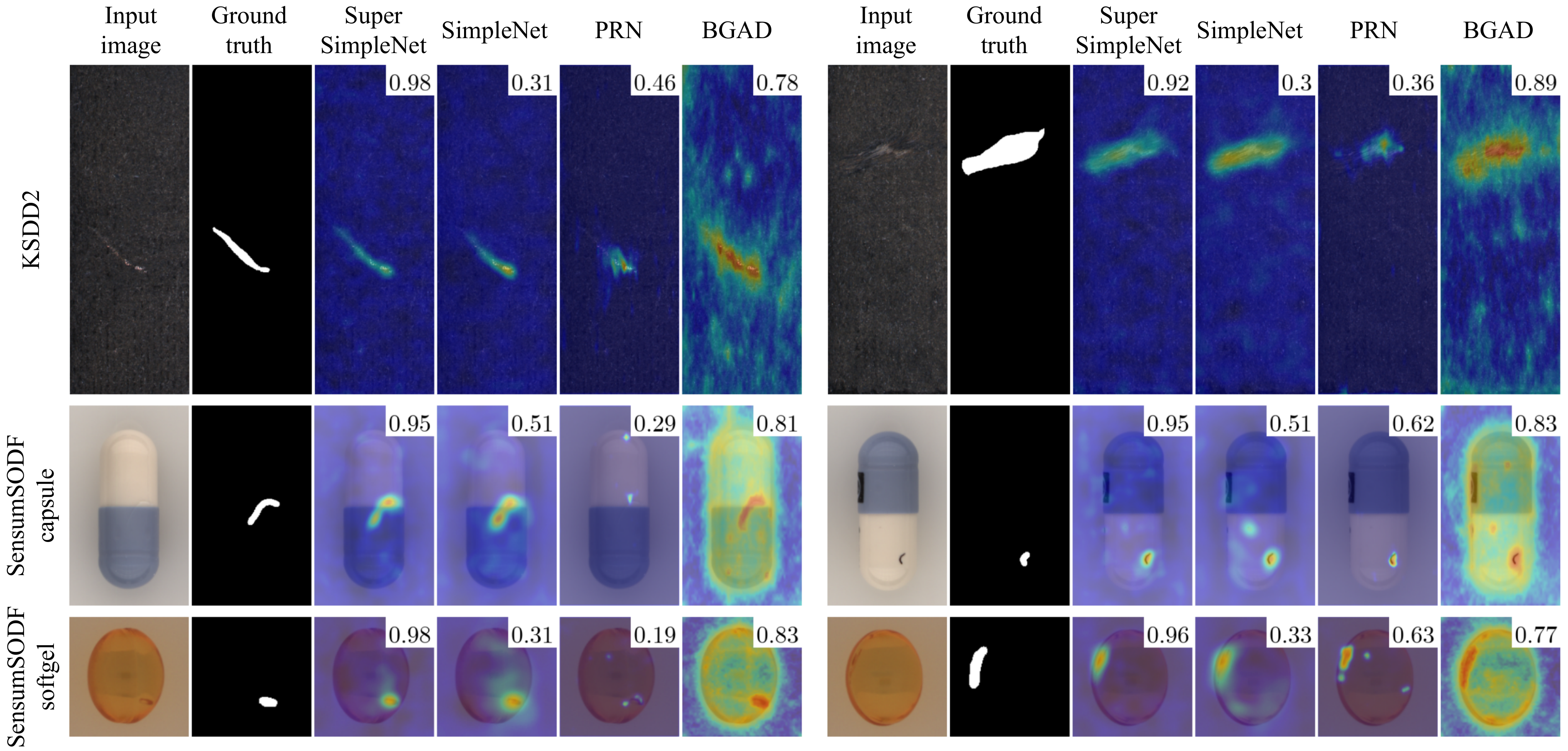}
   \caption{Qualitative comparison of anomaly maps produced in the supervised setting: the input image, the ground truth, and the overlaid anomaly map for SuperSimpleNet, SimpleNet, PRN, and BGAD. The first row displays two samples from KSDD2; the second and third are SensumSODF capsule and softgel respectively. The anomaly score is displayed in the top right corner of each overlaid anomaly map. 
   The anomaly score from the classification head proves to be more reliable than the established maximum value of the anomaly mask.
   }    
   \label{fig:sup_qual}
\end{figure}
\vspace{-2mm}

Figure~\ref{fig:sup_qual} shows qualitative results of our supervised model on KSDD2~\cite{KSDD2} and SensumSODF~\cite{racki_sensum}.
It can be discerned that SuperSimpleNet improves upon previous methods in two aspects: first, it outputs more precise masks in comparison to its competitors PRN~\cite{zhang_prn}, BGAD~\cite{yao_bgad} and its baseline SimpleNet~\cite{liu_simplenet}. PRN outputs masks that majorly underestimate the total area of the defect, whilst BGAD heavily overestimates it. The more precise masks are due to the upscaling module and improved training. SuperSimpleNet improves upon both and reduces the area of the false positive regions produced by the baseline SimpleNet~\cite{liu_simplenet}. Secondly, SuperSimpleNet estimates the image-level anomaly score more accurately than its competitors due to its classification head. While SimpleNet and PRN often predict scores below 0.5, SuperSimpleNet predicts scores near the upper limit.

\noindent\textbf{Results in the unsupervised setting.} We compared our method with the current state-of-the-art methods for the unsupervised setting: AST~\cite{rudolph_ast}, DSR~\cite{zavrtanik2022dsr}, EfficientAD~\cite{batzner_efficientad}, FastFlow~\cite{yu_fastflow}, Patchcore~\cite{roth_patchcore}, DR{\AE}M~\cite{zavrtanik_draem} and SimpleNet~\cite{liu_simplenet}. Table~\ref{tab:mvtec} reports performance on the MVTec AD dataset. 

SuperSimpleNet achieves state-of-the-art performance with mean anomaly detection of $98.4\%$. Results on the VisA dataset are shown in Table~\ref{tab:visa}. SuperSimpleNet reaches state-of-the-art performance with anomaly detection AUROC of $93.4\%$. Whilst the classification head can't learn as efficiently as in the supervised setting, the improved synthetic anomaly generation improves the models' discriminative capabilities, leading to better downstream anomaly detection.
\vspace{-5mm}

\begin{table}[!h]
    \fontsize{7.5pt}{7.5pt}\selectfont
    \setlength{\tabcolsep}{2pt}
    \centering
    \begin{tabular}{lcccccccc}
    \toprule
		~  & \makecell{\cite{rudolph_ast}  \\} & \makecell{\cite{zavrtanik2022dsr}  \\} & \makecell{\cite{batzner_efficientad}  \\} & \makecell{\cite{yu_fastflow}  \\} & \makecell{\cite{roth_patchcore}  \\} & \makecell{\cite{zavrtanik_draem}  \\} & \makecell{\cite{liu_simplenet}  \\} & \makecell{Ours} \\ \midrule
		Carpet & 98.3/89.4& 99.9/92.4& 99.3/92.7& 97.5/92.9& 98.7/93.0& 97.0/92.9& 97.8/92.3& 98.4/92.3\\
		Grid & 98.7/79.7& 100/88.9& 99.9/88.9& 100/96.0& 98.2/91.9& 99.9/98.4& 99.6/93.5& 99.3/93.1\\
		Leather & 100/90.4& 99.6/97.3& 100/98.3& 100/99.1& 100/96.9& 100/97.8& 100/96.0& 100/96.9\\
		Tile & 99.1/72.0& 100/85.6& 100/85.7& 99.9/87.3& 100/87.9& 99.6/98.5& 98.8/85.4& 99.7/84.2\\
		Wood & 99.2/71.9& 92.2/84.5& 99.5/90.2& 98.9/93.1& 99.1/85.7& 99.1/93.5& 97.3/77.7& 99.3/84.7\\ \midrule
		Bottle & 100/86.0& 99.8/95.9& 100/95.7& 100/89.3& 100/94.0& 99.2/97.0& 100/92.9& 100/90.4\\
		Cable & 98.0/75.6& 96.6/88.5& 95.2/92.5& 93.9/89.9& 99.5/94.1& 91.8/75.6& 98.9/90.6& 98.1/88.5\\
		Capsule & 98.8/88.1& 96.7/89.8& 97.9/97.6& 98.1/95.4& 98.5/93.4& 98.5/91.0& 98.0/91.3& 98.7/92.3\\
		Hazelnut & 100/89.5& 99.5/94.7& 99.4/95.7& 98.9/95.6& 100/95.1& 100/98.6& 99.3/89.4& 99.8/94.5\\
		Metal nut & 97.8/75.6& 99.8/91.8& 99.6/94.4& 99.6/92.3& 99.9/94.1& 98.7/94.0& 99.1/88.3& 99.5/90.9\\
		Pill & 99.0/71.7& 98.3/95.9& 98.6/96.1& 96.7/93.9& 95.1/93.9& 98.9/88.2& 97.0/93.3& 98.1/94.2\\
		Screw & 99.1/87.1& 95.8/91.3& 97.0/96.4& 84.5/89.7& 97.3/94.6& 93.9/98.2& 88.7/91.3& 92.9/95.3\\
		Toothbrush & 97.5/67.1& 100/95.8& 100/93.3& 89.2/87.0& 95.3/85.7& 100/90.3& 89.9/91.8& 92.2/85.1\\
		Transistor & 98.9/90.6& 93.5/78.9& 99.9/91.2& 98.5/92.0& 99.8/94.8& 93.1/81.4& 99.5/90.0& 99.9/91.5\\
		Zipper & 99.1/83.2& 99.7/91.1& 99.7/93.4& 98.5/93.7& 99.2/94.7& 100/96.2& 99.4/94.5& 99.6/93.1\\ \midrule
		\textit{Average} & 98.9/81.2& 98.1/90.8& 99.1/93.5& 96.9/92.5& 98.7/92.7& 98.0/92.8& 97.6/90.5& 98.4/91.1\\
        \bottomrule
    \end{tabular}
    \caption{Anomaly detection and localisation (AUROC/AUPRO) on MVTec AD dataset.}
    \label{tab:mvtec}
\end{table}

\begin{table}[!h]
    \fontsize{7.5pt}{7.5pt}\selectfont
    \setlength{\tabcolsep}{2pt}
    \centering
    \begin{tabular}{lcccccccc} 
        \toprule
		~  & \makecell{\cite{rudolph_ast}  \\} & \makecell{\cite{zavrtanik2022dsr}  \\} & \makecell{\cite{batzner_efficientad}  \\} & \makecell{\cite{yu_fastflow}  \\} & \makecell{\cite{roth_patchcore}  \\} & \makecell{\cite{zavrtanik_draem}  \\} & \makecell{\cite{liu_simplenet}  \\} & \makecell{Ours} \\ \midrule
		Candle & 99.4/94.1& 86.4/79.7& 98.4/95.7& 96.8/93.7& 98.6/96.4& 92.7/92.7& 92.5/89.8& 97.1/93.6\\
		Capsules & 85.4/68.2& 93.4/74.5& 93.5/96.9& 83.0/89.3& 76.4/57.5& 90.2/85.4& 78.9/84.8& 81.5/80.1\\
		Cashew & 95.1/79.1& 85.2/61.5& 97.2/94.2& 90.0/84.7& 97.9/88.8& 85.5/67.6& 91.9/82.6& 93.0/86.3\\
		Chewing gum & 100/78.5& 97.2/58.2& 99.9/83.1& 99.8/86.8& 98.9/75.9& 95.2/58.0& 99.0/84.1& 99.3/87.7\\
		Fryum & 99.0/58.7& 93.0/65.5& 96.5/86.7& 98.6/72.9& 94.8/80.8& 88.6/80.4& 95.4/90.3& 96.8/79.3\\
		Macaroni 1 & 93.9/87.2& 91.7/57.7& 99.4/99.0& 94.8/94.1& 95.8/75.0& 94.2/86.3& 94.2/97.3& 93.1/95.9\\
		Macaroni 2 & 72.1/80.4& 79.0/52.2& 96.7/98.8& 80.5/87.3& 77.7/49.1& 86.6/96.3& 71.8/85.9& 75.0/89.7\\
		PCB1 & 99.2/89.3& 89.1/61.3& 98.5/97.1& 95.5/91.2& 98.9/91.3& 75.9/61.1& 92.5/88.7& 96.9/92.9\\
		PCB2 & 98.4/85.7& 96.4/84.9& 99.5/95.0& 96.1/87.0& 97.1/85.7& 98.9/76.2& 93.6/89.0& 97.5/85.4\\
		PCB3 & 97.4/87.7& 97.0/79.5& 98.9/94.0& 94.0/77.6& 96.3/73.2& 94.4/83.5& 92.6/90.3& 94.4/83.0\\
		PCB4 & 99.7/80.1& 98.5/62.1& 98.9/92.7& 98.4/88.6& 99.4/88.7& 98.6/73.4& 97.9/81.6& 98.4/87.3\\
		Pipe fryum & 99.4/89.4& 94.3/80.5& 99.7/94.7& 99.6/89.0& 99.7/94.5& 97.6/74.6& 94.6/91.1& 97.6/88.0\\
        \midrule
		\textit{Average} & 94.9/81.5& 91.8/68.1& 98.1/94.0& 93.9/86.8& 94.3/79.7& 91.5/78.0& 91.2/88.0& 93.4/87.4\\
  \bottomrule
    \end{tabular}
    \caption{Anomaly detection and localisation (AUROC/AUPRO) on VisA dataset.}
    \label{tab:visa}
\end{table}
\vspace{-9mm}

\begin{figure}[!h]
  \centering
   \includegraphics[width=1\linewidth]{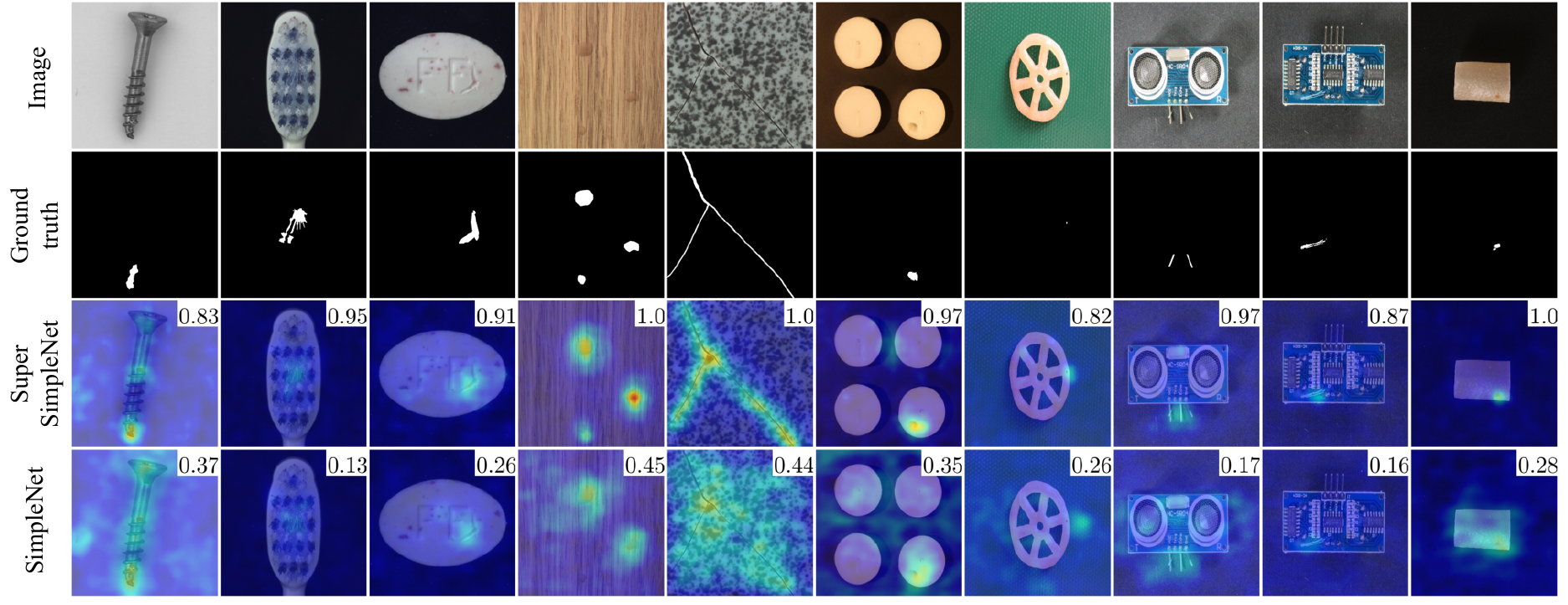}
   \caption{Qualitative comparison of anomaly maps produced by unsupervised SuperSimpleNet and SimpleNet. The top row shows the input anomalous image. The second row displays the ground truth anomaly mask. The third and fourth rows contain anomaly maps generated by SuperSimpleNet and SimpleNet, respectively. The anomaly score is displayed in the top right corner of each anomaly map. 
   }    
   \label{fig:unsup_qual}
\end{figure}

\noindent Qualitative results of the unsupervised model on MVTec AD~\cite{mvtec} and VisA~\cite{visa} are shown in Figure~\ref{fig:unsup_qual}. Like the supervised model, SuperSimpleNet outputs more precise masks than SimpleNet and outputs higher anomaly scores. Similarly, SuperSimpleNet outputs a lower rate of false positives in the background than SimpleNet.

\begin{figure}[!h]
  \centering
   \includegraphics[width=1\linewidth]{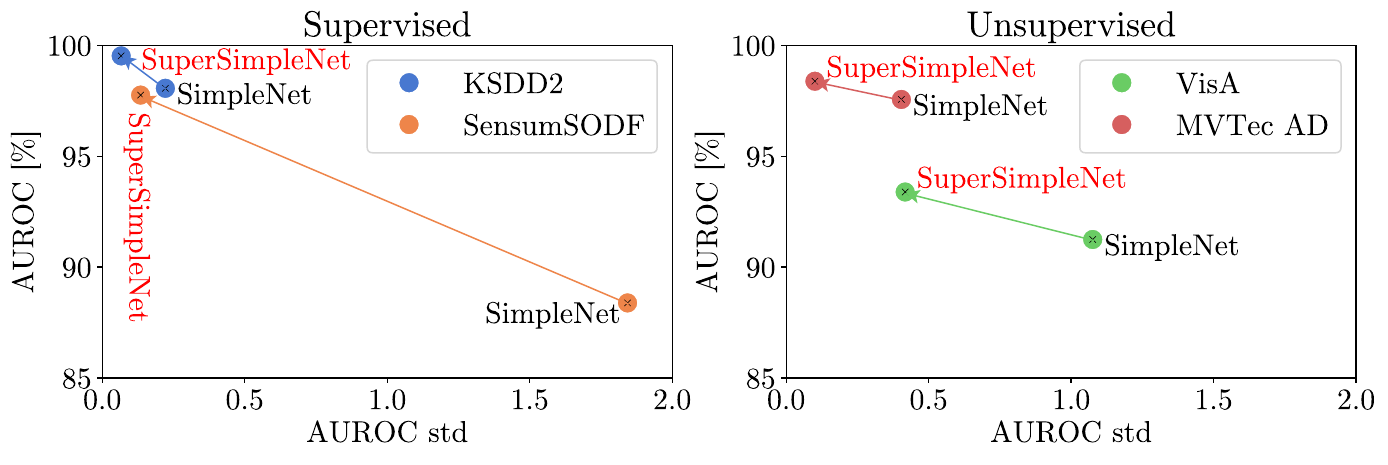}
   \caption{Comparison of SuperSimpleNet with SimpleNet in anomaly detection in terms of AUROC and its standard deviation on all four datasets.}    
   \label{fig:mean_std}
\end{figure}

\noindent\textbf{Stability comparison.} We also evaluated the stability of the training. A comparison in anomaly detection performance and the standard deviation between several training runs with SimpleNet is shown in Figure~\ref{fig:mean_std}. SuperSimpleNet improves the performance of SimpleNet and reduces the standard deviation between several training runs, making it more reliable and robust. The same holds even for the unsupervised setting for which SimpleNet was specifically designed.

\noindent\textbf{Computational efficiency.} Figure~\ref{fig:model_comp} showcases the balance between performance and computational efficiency, measured using an NVIDIA Tesla V100S. SuperSimpleNet is the only model designed to work well in both supervised and unsupervised regimes. At the same time, SuperSimpleNet offers great anomaly detection performance while fulfilling the low inference time requirement with an inference time of 9.3 ms and a throughput of 268 images per second. 
The protocol from~\cite{batzner_efficientad} is followed to measure computational efficiency. A more detailed description and additional metrics are presented in Supplementary Material~\ref{ap:perf}.

\section{Ablation Study}
\label{sec:ablation}

\begin{figure}[!t]
  \centering
   \includegraphics[width=1\linewidth]{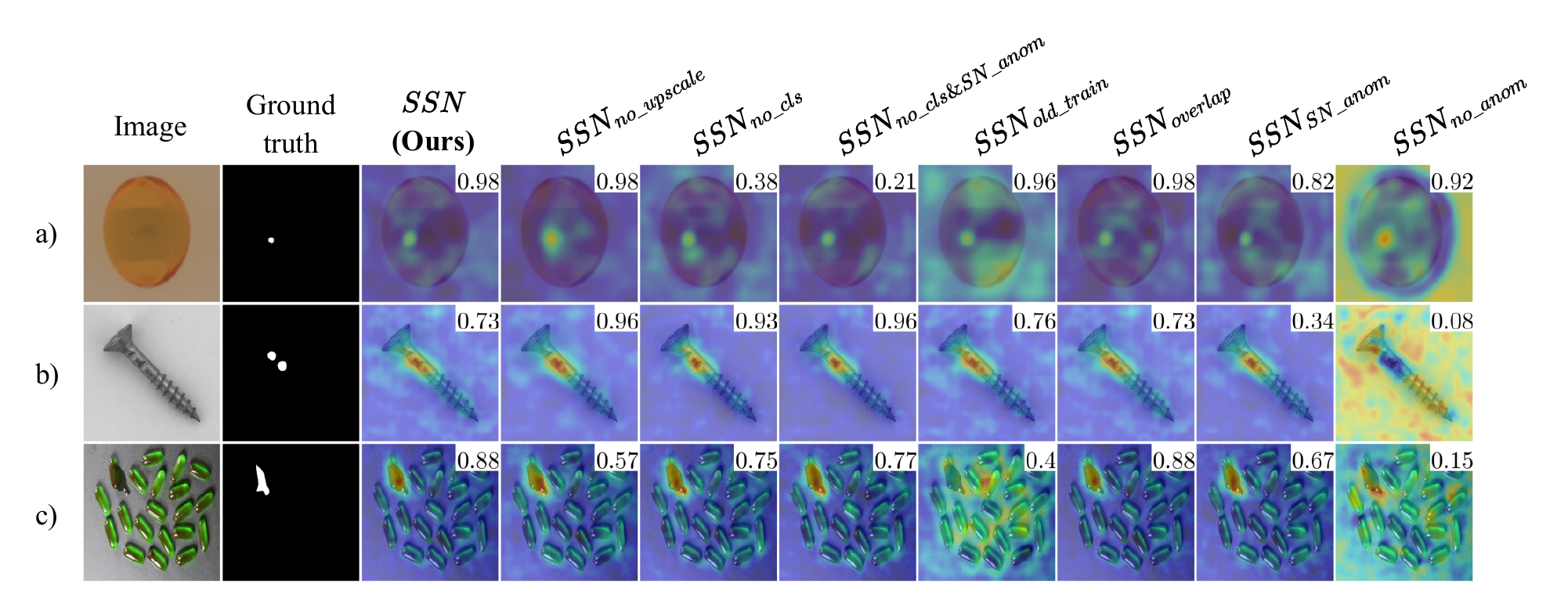}
   \caption{Qualitative comparison of anomaly maps produced by different versions of SuperSimpleNet from ablation study. a) shows a sample from the supervised setting; b) and c) show two samples from the unsupervised setting. The anomaly score is displayed in the top right corner of each overlaid anomaly map. Observing both the anomaly map and the anomaly score provides an insight into how each component contributes towards our final model.
   }    
   \label{fig:ablation_qual}
\end{figure}

To determine the contributions of each component in SuperSimpleNet, each newly introduced module is evaluated by excluding it from the architecture. Quantitative results are shown in Table~\ref{tab:components}, while the qualitative results are shown in Figure~\ref{fig:ablation_qual}. 
\vspace{-5mm}
\begin{table}[!ht]
    \centering
    \setlength{\tabcolsep}{2pt}
    \resizebox{\linewidth}{!}{
        \begin{tabular}{lccccccccccc}
        \toprule
        \multirow{2}{*}{\textbf{Method}}~ &
        \multicolumn{2}{c}{Anomaly generation} &
        \multicolumn{3}{c}{Architecture} &
        \multicolumn{2}{c}{\textit{Super.}} &
        \multicolumn{2}{c}{\textit{Unsup.}} & 
        \multicolumn{2}{c}{\textit{Average}}\\
        
         ~ & SuperSimpleNet & SimpleNet & Upscale & Cls. head & Opt. train. & Det.  & Loc.  & Det.  & Loc. & Det.  & Loc.\\ \midrule
 \rowcolor{gray!20} $SSN$ \textbf{(Ours)} & \checkmark & ~ & \checkmark & \checkmark & \checkmark & 98.6& 95.6& 95.9& 89.3& 97.3& 92.4\\ 

$SSN_{no\_upscale}$ & \checkmark & ~ & ~ & \checkmark & \checkmark &98.2& 93.0& 94.9& 88.3& 96.6& 90.6\\ 
$SSN_{no\_cls}$ & \checkmark & ~ & \checkmark & ~ & \checkmark &95.5& 96.2& 96.2& 89.6& 95.9& 92.9\\ 
$SSN_{no\_cls\&SN\_anom}$ & ~ & \checkmark & \checkmark & ~ & \checkmark & 94.9& 96.6& 96.0& 89.1& 95.4& 92.8\\ 
$SSN_{old\_train}$ & \checkmark & ~ & \checkmark & \checkmark & ~ &98.2& 93.5& 92.7& 85.3& 95.5& 89.4\\ 

$SSN_{overlap}$ & Overlap & ~ & \checkmark & \checkmark & \checkmark& 98.3& 95.9& 95.9& 89.3& 97.1& 92.6\\ 
$SSN_{SN\_anom}$ & ~ & \checkmark & \checkmark & \checkmark & \checkmark &97.9& 96.5& 88.0& 87.5& 93.0& 92.0\\ 
$SSN_{no\_anom}$ & ~ & ~ & \checkmark & \checkmark & \checkmark &98.1& 89.8& - & - & - & - \\ 

\midrule %
 
$SN_{SSN\_anom}$ & \checkmark & ~ & ~ & ~ & ~ &91.3& 91.0& 94.6& 86.2& 93.0& 88.6\\ 
$SN$ & ~ & \checkmark & ~ & ~ & ~ & 93.2& 93.2& 94.4& 89.3& 93.8& 91.2\\ 
\bottomrule
        \end{tabular}
        }
    \caption{Ablation study results on anomaly detection and localisation (AUROC / AUPRO) in the supervised setting (mean value of results on SensumSODF and KSDD2) and the unsupervised setting (mean value of results on MVTec AD and VisA), as well as the average of both settings. $SSN$ stands for SuperSimpleNet, while $SN$ stands for SimpleNet.}
    \label{tab:components}
\end{table}
\vspace{-5mm}

\noindent\textbf{Upscaling module.} Excluding the upscaling of features ($SSN_{no\_upscale}$) causes a decline in detection performance, a 0.4 p.p. decrease for the supervised setting and a 1.0 p.p. in the unsupervised setting. A noticeable decline also occurs in the localisation performance with a drop of performance by 2.6 p.p. and 1.0 p.p. for the supervised setting and the unsupervised setting, respectively. The results suggest the importance of feature size in the final predictions, as also visible from less precise segmentation in Figure~\ref{fig:ablation_qual}.

\noindent\textbf{Classification head.} The exclusion of a classification head ($SSN_{no\_cls}$ - where the anomaly score $s$ is obtained as the maximum value from the anomaly map) notably reduces anomaly detection performance (by 3.1 p.p) in the supervised setting. This is due to stronger discriminative capabilities inside the classification head, which are beneficial when both synthetic and real data are available. The decreased detection performance is also indicated by lower anomaly scores in Figure~\ref{fig:ablation_qual}. However, excluding the classification head in the unsupervised setting leads to slight improvement (by 0.3 p.p). This is because the strong discriminative properties can hinder performance when relying solely on synthetic data. The gradient also flows from the classification head to the segmentation head in the supervised setting, adjusting the anomaly map to be more suitable for classification. Removing the classification head thus slightly increases localisation performance.

\noindent\textbf{Improved training.} The effect of upgrading the loss, incorporating a scheduler, and gradient adjustments can be deduced from the $SSN_{old\_train}$ experiment in the table. These modifications notably impact both detection (by 0.4 p.p. and by 3.2 p.p.) and localisation (by 2.1 p.p. and by 4.0 p.p.). Unsupervised performance is particularly improved, largely due to a more stable training process, which prevents poor final results like the capsules in Figure~\ref{fig:ablation_qual} -- row c. 

\noindent\textbf{Addition of synthetic anomalies to anomalous regions.}
In the supervised setting, synthetic anomalies are exclusively generated within non-anomalous regions. The results of the $SSN_{overlap}$ experiment showcase the impact of anomaly generation without this constraint. This approach leads to slightly worse detection performance (0.3 p.p.). We hypothesise that augmenting already anomalous regions leads to the loss of genuine anomalous information. Since this change only applies to the supervised setting, the unsupervised results remain unaffected. 

\noindent\textbf{Anomaly mask generation.} The importance of anomaly mask generation was evaluated by using the feature duplication strategy from SimpleNet. SimpleNet copies the features and adds the noise to the entirety of the copy. The strategy from SimpleNet ($SSN_{SN\_anom}$) leads to a decrease in detection performance in both the supervised (by 0.7 p.p.) and the unsupervised setting (by 7.9 p.p.). We hypothesise the major decline in unsupervised performance is due to the incompatibility of this mask generation strategy with the classification head, as it struggles to efficiently learn defect bordering regions. This hypothesis is further supported by the $SSN_{no\_cls\&SN\_anom}$ experiment, where SimpleNet strategy is used for SuperSimpleNet with the classification head removed. This improves the unsupervised performance but leads to poor supervised performance, indicating that our anomaly generation strategy is crucial for good simultaneous supervised and unsupervised performance when using a classification head.

\noindent\textbf{Synthetic anomaly generation strategy.} To evaluate the importance of synthetic anomalies in the supervised setting, only real anomalies were used during training. As evident from the $SSN_{no\_anom}$ experiment in the table, the detection performance achieves a decline of 0.5 p.p., whilst the localisation performance achieves a major decline of 5.8 p.p. The results indicate the importance of synthetic anomalies during training. The results from the unsupervised setting are omitted due to the inability of the model to learn a boundary without the presence of synthetic anomalies during training, also seen in Figure~\ref{fig:ablation_qual}.

\section{Conclusion}

A novel discriminative anomaly detection model, SuperSimpleNet, has been proposed to meet the industry's requirements (performance, speed, robustness, and stable training). It offers the flexibility to be trained in both the supervised and the unsupervised setting, making full use of all available training data.
This ability is rarely achieved in previous methods. The proposed model is also robust, achieving a predictable performance independent of the training run. The efficiency of SuperSimpleNet is validated in both the supervised and the unsupervised setting. In the supervised setting, the method is evaluated on two well-established benchmarks, SensumSODF and KSDD2, achieving 97.8\% AUROC on SensumSODF and 97.4\% $\text{AP}_{\text{det}}$ on KSDD2. On SensumSODF, SuperSimpleNet surpasses all previous methods by 0.9\%. SuperSimpleNet also achieves state-of-the-art results in the unsupervised setting on two well-established benchmarks, MVTec AD and VisA, with 98.4\% and 93.4\% AUROC, respectively. It achieves state-of-the-art results whilst holding an inference time of 9.3 ms and a throughput of 268 images per second. 

\noindent\textbf{Limitations and future work.} SuperSimpleNet mostly struggles in the unsupervised setting with categories containing multiple objects. SuperSimpleNet is also heavily dependent on the used backbone, as the magnitude of the Gaussian noise needs to be adjusted for each backbone. Also, anomaly detection performance deteriorates if the backbone fails to extract meaningful features.
In the future, we will extend the model to also operate in weakly-supervised and mixed-supervised settings. This means we will have pixel-level annotations for only a subset of defective images while having image-level annotations for the entire training set. Such a change will further improve direct usability and alleviate the need for pixel-level annotations.  
The results also indicate that combining the knowledge from the unsupervised and the supervised domain is a viable step for the anomaly detection field in the future.

\subsubsection{Acknowledgements} This work was in part supported by the ARIS research project L2-3169 (MV4.0), research programme P2-0214 and the supercomputing network SLING (ARNES, EuroHPC Vega).

%
%
%
\bibliographystyle{splncs04}
\bibliography{egbib}

\clearpage
\setcounter{page}{1}
\clearpage
\setcounter{page}{1}
\maketitlesupplementary

\appendix
\section{Computational efficiency benchmark details}
\label{ap:perf}

\subsection{Implementation details}

The computational efficiency metrics are obtained following the protocol outlined in the EfficientAD paper~\cite{batzner_efficientad}. The models are evaluated on a system with AMD Epyc 7272 CPU and NVIDIA Tesla V100S GPU. The official implementations were used where possible. For PRN, EfficientAD, and FastFlow the unofficial implementations\footnote{\url{https://github.com/xcyao00/PRNet/tree/2cbf85634551fc1b2350cf6de2f80e7854a6c0be}}\footnote{\url{https://github.com/nelson1425/EfficientAD/tree/fcab5146f84ae17597044ad5ddf1656ccf805401}}\footnote{\url{https://github.com/openvinotoolkit/anomalib/tree/4abfa93dcfcb98771bc768b334c929ff9a02ce8b}} were used.

For AST~\cite{rudolph_ast}, DSR~\cite{zavrtanik2022dsr}, FastFlow~\cite{yu_fastflow}, SimpleNet~\cite{liu_simplenet} and PatchCore~\cite{roth_patchcore} the same parameter setup is used as described in~\cite{batzner_efficientad}. For other methods, the default parameters set by the authors are utilised. For SuperSimpleNet, the same parameters are used as described in Section~\ref{exp:impl}.
For SimpleNet, the code provided by authors has redundant feature reshaping in the inference step that significantly impacts the inference time (22ms vs 60ms). That part is omitted for our measurements.

\subsection{Results for all methods}

Table~\ref{ap:tab:perf} contains additional metrics and the numeric values of data presented in Figure~\ref{fig:model_comp} in the main paper. Note that SuperSimpleNet has a total of 34M parameters, but only 9M are trainable, as the rest comes from the frozen pretrained backbone. 

Similarly, as in~\cite{batzner_efficientad}, the number of TFLOPs can be misleading when it comes to measuring efficiency. The GPU memory footprint of each method is mostly stable across the runs, but it can sometimes fluctuate. We hypothesise that this comes from the nondeterministic properties of PyTorch Cuda backend (even when using \verb|torch.backends.cudnn.deterministic = True|).

\begin{table}[!t]
    \centering
    \setlength{\tabcolsep}{4pt}
    \begin{tabular}{lccccc}
    \toprule
    Method & \makecell{Inference \\\relax time [ms]} & \makecell{Throughput \\\relax [img/s]} & \makecell{Num. of \\ Parameters \\\relax [$\times 10^6$]} & \makecell{FLOPs \\\relax [$\times 10^9$]} & \makecell{GPU Memory \\\relax [MB]} \\ \hline
    SegDecNet~\cite{KSDD2}& $4.4 \pm 0.18$& $897.7 \pm 1.98$& 15& 42.4& $134.2 \pm 0.0$\\ 
    DRA~\cite{ding_dra}& $8.3 \pm 0.05$& $1813.6 \pm 21.83$& 14& 35.9& $86.0 \pm 0.0$\\ 
    BGAD~\cite{yao_bgad}& $46.7 \pm 3.08$& $140.9 \pm 3.82$& 39& 14.7& $326.1 \pm 2.57$\\ 
    PRN~\cite{zhang_prn}& $25.4 \pm 2.55$& $128.3 \pm 0.31$& 88& 76.9& $910.2 \pm 0.0$\\ 
    DR{\AE}M~\cite{zavrtanik_draem}& $12.4 \pm 0.02$& $107.3 \pm 0.04$& 97& 395.9& $352.3 \pm 0.0$\\ 
    DSR~\cite{zavrtanik2022dsr}& $18.4 \pm 0.06$& $105.2 \pm 0.14$& 40& 266.5& $511.7 \pm 0.0$\\ 
    PatchCore~\cite{roth_patchcore}& $54.1 \pm 1.36$& $25.3 \pm 0.64$& 128& 53.7& $467.7 \pm 0.0$\\ 
    AST~\cite{rudolph_ast}& $57.9 \pm 0.55$& $257.0 \pm 3.62$& 145& 20.9& $521.4 \pm 22.76$\\ 
    FastFlow~\cite{yu_fastflow}& $21.3 \pm 0.03$& $320.4 \pm 0.18$& 92& 85.3& $432.0 \pm 0.0$\\ 
    SimpleNet~\cite{liu_simplenet}& $22.4 \pm 2.42$& $95.2 \pm 3.1$& 73& 37.8& $260.0 \pm 0.0$\\ 
    EfficientAD~\cite{batzner_efficientad}& $6.0 \pm 0.01$& $204.0 \pm 0.08$& 21& 235.4& $178.3 \pm 0.0$\\ \midrule
    SuperSimpleNet& $9.3 \pm 0.1$& $268.8 \pm 0.22$& 34& 56.3& $361.5 \pm 27.13$\\ 
    \bottomrule
    \end{tabular}
    \caption{Computational efficiency metrics data, measured on an NVIDIA Tesla V100S. For every model, the average of 5 runs is reported, and the standard deviation where applicable. }
    \label{ap:tab:perf}
\end{table}

\end{document}